\documentclass[10pt,twocolumn,letterpaper]{article}

\usepackage{cvpr}
\usepackage{times}
\usepackage{epsfig}
\usepackage{graphicx}
\usepackage{amsmath}
\usepackage{amssymb}
\usepackage{subfigure}
\usepackage{hhline}

\newcommand{\norm}[1]{\left\lVert#1\right\rVert}

\usepackage[breaklinks=true,bookmarks=false]{hyperref}

\cvprfinalcopy 

\def\cvprPaperID{****} 

\begin{document}

\title{Unsupervised Domain-Specific Deblurring via Disentangled Representations}

\author{Boyu Lu \quad\qquad\qquad Jun-Cheng Chen \qquad\qquad Rama Chellappa\\
UMIACS, University of Maryland, College Park\\
{\tt\small \hspace{-5mm} bylu@umiacs.umd.edu \quad pullpull@cs.umd.edu \quad rama@umiacs.umd.edu}
}

\maketitle

\begin{abstract}
Image deblurring aims to restore the latent sharp images from the corresponding blurred ones.  In this paper, we present an unsupervised method for domain-specific single-image deblurring based on disentangled representations. The disentanglement is achieved by splitting the content and blur features in a blurred image using content encoders and blur encoders. We enforce a KL divergence loss to regularize the distribution range of extracted blur attributes such that little content information is contained. Meanwhile, to handle the unpaired training data, a blurring branch and the cycle-consistency loss are added to guarantee that the content structures of the deblurred results match the original images. We also add an adversarial loss on deblurred results to generate visually realistic images and a perceptual loss to further mitigate the artifacts. We perform extensive experiments on the tasks of face and text deblurring using both synthetic datasets and real images, and achieve improved results compared to recent state-of-the-art deblurring methods.
\end{abstract}

\section{Introduction}

Image blur is an important factor that adversely affects the quality of images and thus significantly degrades the performances of many computer vision applications, such as object detection~\cite{DeblurGAN} and face recognition~\cite{lu2019experimental,lu2015incremental}. To address this problem, blind image deblurring aims to restore the latent sharp image from a blurred image. Most conventional methods formulate the image deblurring task as a blur kernel estimation problem. Since this problem is highly ill-posed, many priors have been proposed to model the images and kernels~\cite{pan2016blind,xu2013unnatural,kingma2013auto}. However, most of these priors only perform well on generic natural images, but cannot generalize to specific image domains, like face~\cite{shen2018deep}, text~\cite{BMVC2015_6} and low-illumination images~\cite{hu2014deblurring}. Therefore, some priors (\eg $L_0$-regularized intensity and gradient prior~\cite{pan2014deblurring}, face exemplars~\cite{pan2014deblurring2}) have been developed to handle these domain-specific image deblurring problems. But these methods still can only handle certain types of blur and often require expensive inference time.

Recently, some learning-based approaches have been proposed for blind image deblurring~\cite{DeblurGAN,nah2017deep,shen2018deep}. CNN-based models can handle more complex blur types and have enough capacity to train on large-scale datasets. These models can be trained end-to-end and the inference is fast due to GPU acceleration. Meanwhile, the Generative Adversarial Networks (GAN) have been found to be effective in generating more realistic images. Nonetheless, most of these methods need paired training data, which is expensive to collect in practice. Although numerous blur generation methods have been developed~\cite{DeblurGAN,sun2015learning,chakrabarti2016neural}, they are not capable of learning all types of blur variants in the wild. Moreover, strong supervision may cause algorithms to overfit training data and thus cannot generalize well to real images. 

\begin{figure}
    \begin{center}
    \subfigure[Blurred]{ 
        \includegraphics[width=0.31\linewidth]{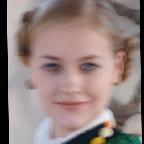}
    }\hspace*{-2mm}
    \subfigure[Madam \etal~\cite{madam2018unsupervised}]{
    \label{fig:Nimisha}
        \includegraphics[width=0.31\linewidth]{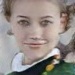}
    }\hspace*{-2mm}
    \subfigure[Ours]{
        \includegraphics[width=0.31\linewidth]{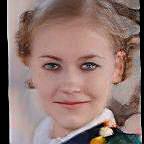}
    }
    \subfigure[Blurred]{
        \includegraphics[width=0.31\linewidth]{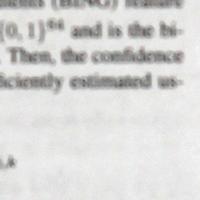}
    }\hspace*{-2mm}
    \subfigure[CycleGAN~\cite{CycleGAN2017}]{
    \label{fig:intro_cyclegan_text}
        \includegraphics[width=0.31\linewidth]{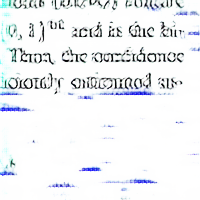}
    }\hspace*{-2mm}
    \subfigure[Ours]{
        \includegraphics[width=0.31\linewidth]{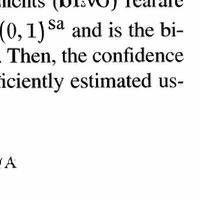}
    }
    \caption{Qualitative deblurred results of the proposed method compared with other state-of-the-art unpaired deblurring methods on real-world blurred face and text images. The deblurred image of (b) is from~\cite{madam2018unsupervised}. For (e), we apply our trained model using the publicly available code of~\cite{CycleGAN2017}.}
    \label{fig:intro}
    \end{center}
\end{figure}

\begin{figure*}
\begin{center}
\includegraphics[width=0.95\textwidth]{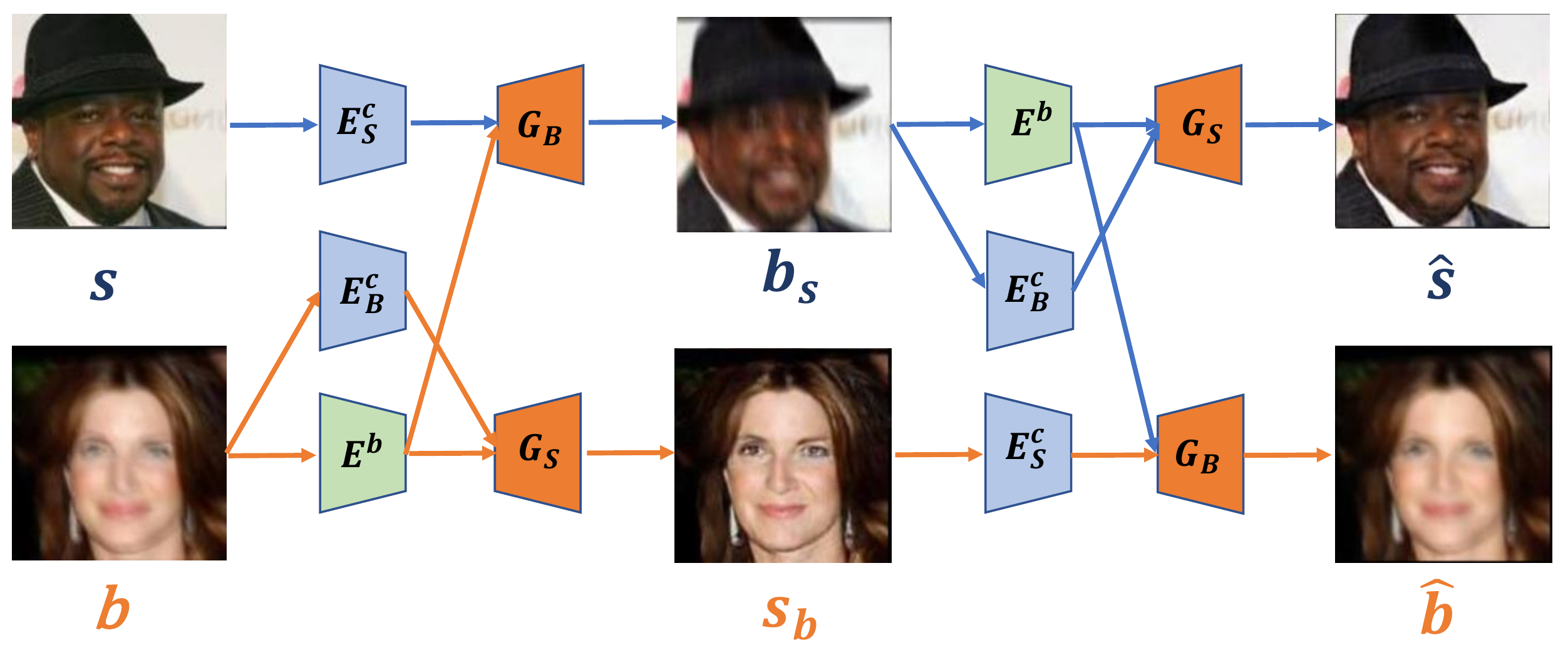}
\end{center}
   \caption{Overview of the deblurring framework.The data flow of the top \textit{blurring} branch (bottom \textit{deblurring} branch) is represented by blue (orange) arrows. 
   $E_B^c$ and $E_S^c$ are content encoders for blurred and sharp images; $E^b$ is blur encoder; $G_B$ and $G_S$ are blurred image and sharp image generators. Two GAN losses are added to distinguish $b_s$ from blur images, and to distinguish $s_b$ from sharp images. The KL divergence loss is added to the output of $E^b$. Cycle-consistency loss is added to $s$ and $\hat{s}$, $b$ and $\hat{b}$. Perceptual loss is added to $b$ and $s_b$.}
\label{fig:system}
\end{figure*}

More recently, Madam Nimisha \etal~\cite{madam2018unsupervised} proposed an unsupervised image deblurring method based on GANs where they add reblur loss and multi-scale gradient loss on the model. Although they achieved good performance on the synthetic datasets, their results on some real blurred images are not satisfactory (Fig.~\ref{fig:Nimisha}). Another solution might be directly using some existing unsupervised methods (CycleGAN~\cite{CycleGAN2017}, DualGAN~\cite{yi2017dualgan}) to learn the mappings between sharp and blurred image domains. However, these generic approaches often encode other factors (\eg, color, texture) rather than blur information into the generators, and thus do not produce good deblurred images (Fig.~\ref{fig:intro_cyclegan_text}).

In this paper, we propose an unsupervised domain-specific image deblurring method based on disentangled representations. More specifically, we disentangle the content and blur features from blurred images to accurately encode blur information into the deblurring framework. As shown in Fig.~\ref{fig:system}, the content encoders extract content features from unpaired sharp and blurred images, and the blur encoder captures blur information. In addition, we share the weights of the last layer of both content encoders so that the content encoders can project the content features of both domains onto a common space. However, this structure by itself does not guarantee that the blur encoder captures blur features - it may encode content features as well. Inspired by~\cite{bao2018towards}, we add a KL divergence loss to regularize the distribution of blur features to suppress the contained content information. Then, the deblurring generator $G_S$ and the blurring generator $G_B$ takes corresponding content features conditioned on blur attributes to generate deblurred and blurred images. Similar to CycleGAN~\cite{CycleGAN2017}, we also use the adversarial loss and the cycle-consistency loss as regularizers to assist the generator networks to yield more realistic images, and also preserve the content of the original image. To further remove the unpleasant artifacts introduced by deblurring generator $G_S$, we add the perceptual loss to the proposed framework. Some sample deblurred images are shown in Fig.~\ref{fig:intro}.

We conduct extensive experiments on face and text deblurring and achieve competitive performance compared with other state-of-the-art deblurring methods. We also evaluate the proposed method on face verification and optical character recognition (OCR) tasks to demonstrate the effectiveness of our algorithm on recovering semantic information. 


\section{Related Works}
Since the proposed approach leverages the recent development of disentangled representations for image deblurring, in this section we briefly review the related works of image deblurring and disentangled representations.
\subsection{Single Image Blind Deblurring}

\textbf{Generic methods} Single image blind deblurring is a highly ill-posed problem. Over the past decade, various natural image and kernel priors have been developed to regularize the solution space of the latent sharp images, including heavy-tailed gradient prior~\cite{shan2008high}, sparse kernel prior~\cite{fergus2006removing}, $l_0$ gradient prior~\cite{xu2013unnatural}, normalized sparsity prior~\cite{krishnan2011blind}, and dark channels~\cite{pan2016blind}. However, these priors are estimated from limited observations, and are not accurate enough. As a result, the deblurred images are often under-deblurred (images are still blurred) or over-deblurred (images contain many artifacts).  

On the other hand, due to the recent immense success of deep networks and GANs, several CNN-based methods have been proposed for image deblurring. Sun \etal~\cite{sun2015learning} and Schuler \etal~\cite{schuler2016learning} use CNN to predict the motion blur kernels. Chakrabarti \etal~\cite{chakrabarti2016neural} predict the Fourier coefficients of the deconvolution filters by a neural network and perform deblurring in frequency domains.  These methods combine the advantage of CNN and conventional maximum a posteriori probability (MAP)-based algorithms. Differently, Nah \etal~\cite{nah2017deep} train a multi-scale CNN in an end-to-end manner to directly deblur images without explicitly estimating the blur kernel. Similarly, Kupyn \etal~\cite{DeblurGAN} use WGAN and perceptual loss and achieved state-of-the-art performance on natural image deblurring.



\textbf{Domain specific methods} Although the above mentioned methods perform well for natural image deblurring, it is difficult to generalize them to some specific image domains, such as face and text images. Pan \etal~\cite{pan2014deblurring} propose the $L_0$-regularized prior on image intensity and gradients for text image deblurring. Hradis \etal~\cite{BMVC2015_6} train an end-to-end CNN specific for text image deblurring. Pan \etal~\cite{pan2014deblurring2} utilize exemplar faces in a reference set to guide the blur kernel estimation. Shen \etal~\cite{shen2018deep} use the face parsing labels as global semantic priors and local structure regularization to improve face deblurring performance.  


\subsection{Disentangled Representation} There has been many recent efforts on learning disentangled representations. Tran \etal~\cite{tran_drgan} propose DR-GAN to disentangle the pose and identity components for pose-invariant face recognition. Bao \etal~\cite{bao2018towards} explicitly disentangle identity features and attributes to learn an open-set face synthesizing model. Liu \etal~\cite{liu2018exploring} construct an identity distill and dispelling auto encoder to disentangle identity with other attributes. BicycleGAN~\cite{zhu2017toward} combines cVAE-GAN and cLR-GAN to model the distribution of possible outputs in image-to-image translation. Recently, some unsupervised methods decouple images into domain-invariant content features and domain-specific attribute vectors, which produce diverse image-to-image translation outputs~\cite{lee2018diverse,almahairi2018augmented,huang2018munit}.

\section{Proposed Method}
The proposed framework consists of four parts: 1) content encoders $E_B^c$ and $E_S^c$ for blurred and sharp image domains; 2) blur encoder $E^b$; 3) blurred and sharp image generators $G_B$ and $G_S$; 4) blurred and sharp image discriminators $D_B$ and $D_S$. Given a training sample $b\in B$ in the blurred image domain and $s\in S$ in the sharp image domain, the content encoders $E_B^c$ and $E_S^c$ extract content information from corresponding samples and $E^b$ estimates the blur information from $b$. $G_S$ then takes $E_B^c(b)$ and $E^b(b)$ to generate a sharp image $s_b$ while $G_B$ takes $E_S^c(s)$ and $E^b(b)$ to generate a blurred image $b_s$. The discriminators $D_B$ and $D_S$ distinguish between the real and generated examples. The end-to-end architecture is illustrated in Fig.~\ref{fig:system}. 

In the following subsections, we first introduce the method to disentangle content and blur components in Section~\ref{disentangle_cb}. Then, we discuss thee loss functions used in our approach. In Section~\ref{testing}, we describe the testing procedure of the proposed framework. Finally, the implementation details are discussed in Section~\ref{exp_details}.

\subsection{Disentanglement of Content and Blur}
\label{disentangle_cb}
Since the ground truth sharp images are not available in the unpaired setting, it is not trivial to disentangle the content information from a blurred image. However, since sharp images only contain content components without any blur information, the content encoder $E_S^c$ should be a good content extractor. We enforce the last layer of $E_B^c$ and $E_S^c$ to share weights so as to guide $E_B^c$ to learn how to effectively extract content information from blurred images. 

On the other hand, the blur encoder $E^b$ should only encode blur information. To achieve this goal, we propose two methods to help $E^b$ suppress as much content information as possible. First, we feed $E^b(b)$ together with $E_S^c(s)$ into $G_B$ to generate $b_s$. Since $b_s$ is a blurred version of $s$ and it will not contain content information of $b$, this structure discourages $E^b(b)$ to encode content information of $b$. Second, we add a KL divergence loss to regularize the distribution of the blur features $z_b = E^b(b)$ to be close to the normal distribution $p(z) \sim N(0, 1)$. As shown in~\cite{bao2018towards}, this will further suppress the content information contained in $z_b$. The KL divergence loss is defined as follows:

\begin{equation}
KL(q(z_b)||p(z)) = -\int q(z_b) \log \frac{p(z)}{q(z_b)} dz
\end{equation}

\noindent{As proved in~\cite{kingma2013auto}, minimizing the KL divergence is equivalent to minimizing the following loss:}
\begin{equation}
\label{KL_loss}
\mathcal{L}_{KL} = \frac{1}{2}\sum_{i=1}^N (\mu^2_i + \sigma_i^2 - \log(\sigma_i^2) - 1)
\end{equation}

\noindent{where $\mu$ and $\sigma$ are the mean and standard deviation of $z_b$ and $N$ is the dimension of $z_b$. Similar to~\cite{kingma2013auto}, $z_b$ is sampled as $z_b = \mu + z \circ \sigma$, where $p(z) \sim N(0, 1)$ and $\circ$ represents element-wise multiplication.}


\subsection{Adversarial Loss}
In order to make the generated images look more realistic, we apply the adversarial loss on both domains. For the sharp image domain, we define the adversarial loss as:
\vspace{-1mm}

\begin{equation}
\label{adv_loss}
\begin{split}
\mathcal{L}_{D_S} =& \quad\mathbb{E}_{s \sim p(s)}[\log D_S(s)] \\
&+ \mathbb{E}_{b \sim p(b)}[\log (1-D_S(G_S(E^c_B(b), z_b)))]
\end{split}
\end{equation}

\noindent{where $D_S$ tries to maximize the objective function to distinguish between deblurred images and real sharp images. In contrast, $G_S$ aims to minimize the loss to make deblurred images look similar to real samples in domain $S$. Similarly, we define the adversarial loss in blurred image domain as $\mathcal{L}_{D_B}$}:
\vspace{-1mm}
\begin{equation}
\label{d_b_loss}
\begin{split}
\mathcal{L}_{D_B} =& \quad\mathbb{E}_{b \sim p(b)}[\log D_B(b)] \\
&+ \mathbb{E}_{s \sim p(s)}[\log (1-D_B(G_B(E^c_S(s), z_b)))]
\end{split}
\end{equation}

\subsection{Cycle-Consistency Loss}
After competing with discriminator $D_S$ in the minmax game, $G_S$ should be able to generate visually realistic sharp images. However, since no pairwise supervision is provided, the deblurred image may not retain the content information in the original blurred image. Inspired by CycleGAN~\cite{CycleGAN2017}, we introduce the cycle-consistency loss to guarantee that the deblurred image $s_b$ can be reblurred to reconstruct the original blurred sample, and $b_s$ can be translated back to the original sharp image domain. The cycle-consistency loss further limits the space of the generated samples and preserve the content of original images. More specifically, we perform the forward translation as:
\vspace{-1mm}
\begin{equation}
\label{forward}
s_b = G_S(E^c_B(b), E^b(b)), b_s = G_B(E^c_S(s), E^b(b))
\end{equation}

\noindent{and the backward translation as:}
\vspace{-0.7mm}
\begin{equation}
\label{backward}
\hat{b} = G_B(E^c_S(s_b), E^b(b_s)), \hat{s} = G_S(E^c_B(b_s), E^b(b_s))
\end{equation}

\noindent{We define the cycle-consistency loss on both domains as:}
\begin{equation}
\label{cc_loss}
\mathcal{L}_{cc} = \mathbb{E}_{s \sim p(s)}[\norm{s - \hat{s}}_1] + \mathbb{E}_{b \sim p(b)}[||b - \hat{b}||_1]
\end{equation}


\subsection{Perceptual Loss}
\label{p_loss}

From our preliminary experiments, we find that the generated deblurred samples often contain many unpleasant artifacts. Motivated by the observations from~\cite{taigman2016unsupervised,chen2017photographic} that features extracted from pre-trained deep networks contain rich semantic information, and their distances can act as perceptual similarity judgments, we add a perceptual loss between the deblurred images and the corresponding original blurred images:
\begin{equation}
\label{p_loss}
\mathcal{L}_{p} = \norm{\phi_l(s_b) - \phi_l(b)}^2_2
\end{equation} 

\noindent{where $\phi_l(x)$ is the features of the $l$-th layer of the pre-trained CNN. In our experiments, we use the $\texttt{conv}_{3,3}$ layer of VGG-19 network~\cite{simonyan2014very} pre-trained on ImageNet~\cite{imagenet_cvpr09}.}

There are two main reasons why we use the blurred image $b$ instead of the sharp one $s$ as the reference image in the perceptual loss. First, we made an assumption that the content information of $b$ can be extracted by the pre-trained CNN. As shown in Fig.~\ref{ablation}, the experimental results confirm this point. Second, since $s$ and $b$ are unpaired, applying the perceptual loss between $s$ and $s_b$ will force $s_b$ to encode irrelevant content information from $s$. However, since we also notice that the perceptual loss is sensitive to blur as shown in~\cite{zhang2018perceptual}, we carefully balance the weight of the perceptual loss with other losses to prevent $s_b$ from staying too close to $b$. The sensitivity evaluation of varying the weight is shown in the supplementary materials.

It is worth mentioning that the perceptual loss is not added on $b_s$ and $s$. This is because we do not find obvious artifacts in $b_s$ during training. Moreover, for text image deblurring, since we observe the percetual loss does not help but sometimes hurt the performance,  we do not include it for this task. One possible reason may be due to the pixel intensity distribution of the text images being very different from the natural images in the ImageNet dataset. 

The full objective function is a weighted sum of all the losses from~(\ref{KL_loss}) to~(\ref{p_loss}):
\vspace{-0.7mm}
\begin{equation}
\label{overall_loss}
\mathcal{L} = \lambda_{adv}\mathcal{L}_{adv} + \lambda_{KL}\mathcal{L}_{KL} + \lambda_{cc}\mathcal{L}_{cc} + \lambda_{p}\mathcal{L}_{p}
\end{equation}

\noindent{where $\mathcal{L}_{adv} = \mathcal{L}_{D_S} + \mathcal{L}_{D_B}$. We empirically set the weights of each loss to balance their importances.} 

\subsection{Testing}
\label{testing}
At test time, the blurring branch is removed. Given a test blurred image $b_t$, $E^c_B$ and $E^b$ extract the content and blur features. Then $G_S$ takes the outputs and generates the deblurred image $s_{b_t}$:
\vspace{-0.7mm}
\begin{equation}
\label{test}
s_{b_t} = G_S(E^c_B(b_t), E^b(b_t))
\end{equation}

\begin{figure*}
\begin{center}
\includegraphics[width=0.95\textwidth]{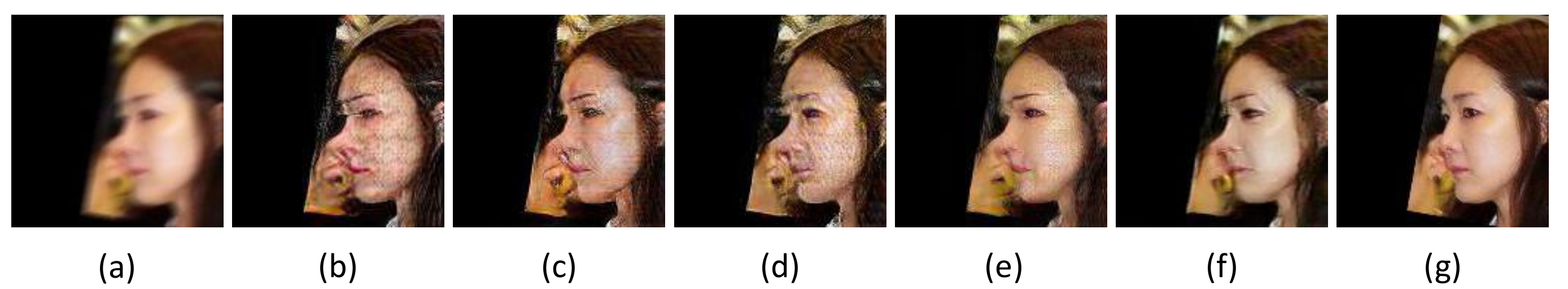}
\end{center}
   \caption{Ablation study. (a) shows the blurred image and (g) is the sharp image. (b) only contains deblurring branch (bottom branch of Fig.~\ref{fig:system}), (c) adds blurring branch (bottom branch of Fig.~\ref{fig:system}), (d) adds disentanglement ($E^b$), (e) adds the KL divergence loss, and (f) adds perceptual loss. } 
\label{fig:ablation}
\end{figure*}

\begin{figure*}
    \hspace*{-3.5mm}
    \subfigure[Blurred]{ \begin{tabular}[]{c}
        \includegraphics[width=0.08\linewidth]{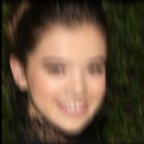} \\
        \includegraphics[width=0.08\linewidth]{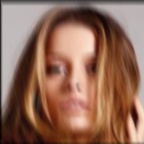}
            \end{tabular}
    }\hspace*{-5.1mm}
    \subfigure[\cite{pan2014deblurring}]{\begin{tabular}[]{c}
            \includegraphics[width=0.081\linewidth]{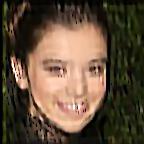} \\
            \includegraphics[width=0.08\linewidth]{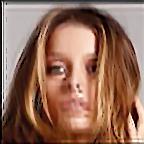}
            \end{tabular}
    }\hspace*{-5.4mm}
    \subfigure[\cite{pan2016blind}]{\begin{tabular}[]{c}
            \includegraphics[width=0.081\linewidth]{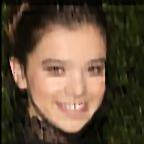} \\
            \includegraphics[width=0.08\linewidth]{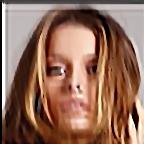}
            \end{tabular}
    }\hspace*{-5.2mm}
    \subfigure[\cite{shen2018deep}]{\begin{tabular}[]{c}
            \includegraphics[width=0.081\linewidth]{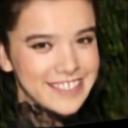} \\
            \includegraphics[width=0.08\linewidth]{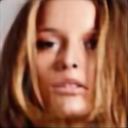}
            \end{tabular}
    }\hspace*{-6.2mm}
    \subfigure[\cite{pan2014deblurring2}]{
        \begin{tabular}[]{c}
            \includegraphics[width=0.081\linewidth]{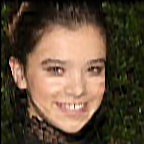} \\
            \includegraphics[width=0.08\linewidth]{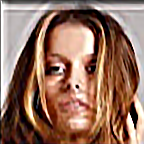}
        \end{tabular}
    }\hspace*{-6mm}
    \subfigure[\cite{xu2013unnatural}]{
        \begin{tabular}[]{c}
            \includegraphics[width=0.081\linewidth]{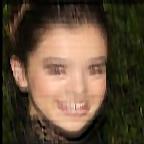} \\
            \includegraphics[width=0.08\linewidth]{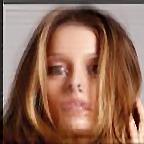}
        \end{tabular}
    }\hspace*{-6.2mm}
    \subfigure[\cite{krishnan2011blind}]{
             \begin{tabular}[]{c}
            \includegraphics[width=0.081\linewidth]{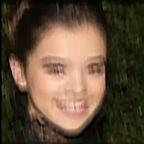} \\
            \includegraphics[width=0.08\linewidth]{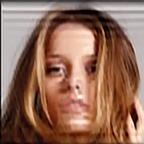}
        \end{tabular}
    }\hspace*{-6.2mm}
    \subfigure[\cite{DeblurGAN}]{
             \begin{tabular}[]{c}
            \includegraphics[width=0.081\linewidth]{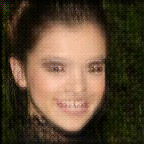} \\
            \includegraphics[width=0.08\linewidth]{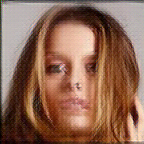}
        \end{tabular}
    }\hspace*{-6.2mm}
    \subfigure[\cite{nah2017deep}]{
            \begin{tabular}[]{c}
            \includegraphics[width=0.081\linewidth]{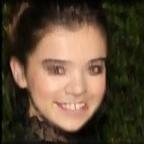} \\
            \includegraphics[width=0.08\linewidth]{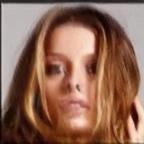}
        \end{tabular}
    }\hspace*{-6mm}
    \subfigure[\cite{CycleGAN2017}]{
             \begin{tabular}[]{c}
            \includegraphics[width=0.081\linewidth]{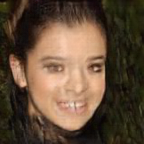} \\
            \includegraphics[width=0.08\linewidth]{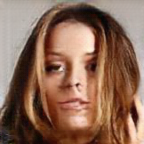}
        \end{tabular}
    }\hspace*{-6.2mm}
    \subfigure[Ours]{
             \begin{tabular}[]{c}
            \includegraphics[width=0.081\linewidth]{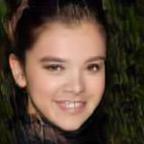} \\
            \includegraphics[width=0.08\linewidth]{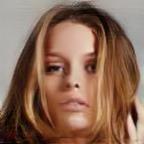}
        \end{tabular}
    }\hspace*{-6.2mm}
    \subfigure[Sharp]{
             \begin{tabular}[]{c}
            \includegraphics[width=0.081\linewidth]{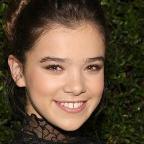} \\
            \includegraphics[width=0.08\linewidth]{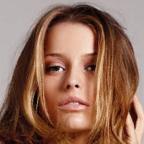}
        \end{tabular}
    }
    \caption{Visual performance comparison with state-of-the-art methods on CelebA dataset. Best viewed in color and by zooming in.}
    \label{fig:celeba}
\end{figure*}

\subsection{Implementation Details}
\label{exp_details}
\textbf{Architecture and training details.} For the network architectures, we follow the similar structures as the one used in~\cite{lee2018diverse}. The content encoder is composed of three strided convolution layers and four residual blocks. The blur encoder contains four strided convolution layers and a fully connected layer. For the generator, the architecture is symmetric to the content encoder with four residual blocks followed by three transposed convolution layers. The discriminator applies a multi-scale structure where feature maps at each scale go through five convolution layers and then are fed into sigmoid outputs. The end-to-end design is implemented in PyTorch~\cite{paszke2017automatic}. During training, we use Adam solver~\cite{kingma2014adam} to perform two steps of update on discriminators, and then one step on encoders and generators. The learning rate is initially set to 0.0002 for the first 40 epochs, then we use exponential decay over the next 40 epochs. In all the experiments, we randomly crop $128\times128$ patches with batch size of 16 for training. For hyper-parameters, we experimentally set: $\lambda_{adv}=1$, $\lambda_{KL}=0.01$, $\lambda_{cc}=10$ and $\lambda_{p}=0.1$. 

\textbf{Motion blur generation.} We follow the procedure in DeblurGAN~\cite{DeblurGAN} to generate motion blur kernels to blur face images. A random trajectory is generated as described in~\cite{boracchi2012modeling}. Then the kernels are generated by applying sub-pixel interpolation to the trajectory vector. For parameters, we use the same values as in~\cite{DeblurGAN} except that we set the probability of impulsive shake as 0.005, the probability of Gaussian shake uniformly distributed in $(0.5,1.0)$, and the max length of the movement as 10.

\begin{table}
\begin{center}
\begin{tabular}{l|c|c|r}
\hline
Method & PSNR & SSIM &$d_{VGG}$\\
\hline
Only deblurring branch & 18.83 & 0.56 & 82.9 \\
Add blurring branch & 19.84  & 0.59  & 65.5 \\
Add disentanglement & 19.58 & 0.57 & 69.8\\
Add KL divergence loss & 20.29 & 0.61 & 60.6 \\
Add perceptual loss & \textbf{20.81} & \textbf{0.65} & \textbf{57.6} \\
\hline
\end{tabular}
\end{center}
\caption{Ablation study on the effectiveness of different components. $d_{VGG}$ represents the distance of feature from VGG-Face, lower is better.}
\label{tab:ablation}
\end{table}

\section{Experimental Results}
We evaluate the proposed approach on three datasets: CelebA dataset~\cite{liu2015faceattributes}, BMVC\_Text dataset~\cite{BMVC2015_6}, and CFP dataset~\cite{cfp-paper}.



\subsection{Datasets and Metrics}
\label{datasets}

\textbf{CelebA dataset:} This dataset consists of more than 202,000 face images. Most of the faces are of good quality and at near-frontal poses. We randomly split the whole dataset into three mutually exclusive subsets: sharp training set (100K images), blurred training set (100K images) and test set (2137 images). For the blurred training set, we use the method in Section~\ref{exp_details} to blur the images. The faces are detected and aligned using the method proposed in~\cite{ranjan2017all}.

\textbf{BMVC\_text dataset:} This dataset is composed of 66,000 text images with size $300\times300$ for training and 94 images with size $512\times512$ for OCR testing. Similar to CelebA, we evenly split the training sets as sharp and blurred set. Since the dataset already contains the blurred text images, we directly use them instead of generating new ones. 

\begin{figure*}
    \begin{center}
    \hspace*{-3.1mm}
    \subfigure[Blurred]{ \begin{tabular}[]{c}
        \includegraphics[width=0.083\linewidth]{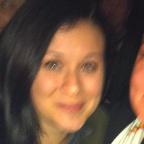} \\
        \includegraphics[width=0.083\linewidth]{figs/real_face/2_blur.jpg}\\
        \includegraphics[width=0.083\linewidth]{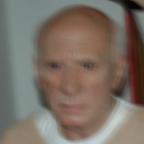}
            \end{tabular}
    }\hspace*{-5.1mm}
    \subfigure[\cite{pan2014deblurring}]{\begin{tabular}[]{c}
            \includegraphics[width=0.083\linewidth]{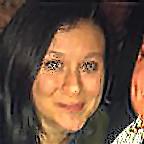} \\
            \includegraphics[width=0.083\linewidth]{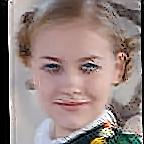} \\
            \includegraphics[width=0.083\linewidth]{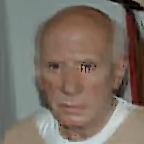}
            \end{tabular}
    }\hspace*{-5.4mm}
    \subfigure[\cite{pan2016blind}]{\begin{tabular}[]{c}
            \includegraphics[width=0.083\linewidth]{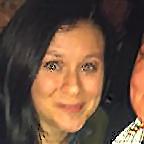} \\
            \includegraphics[width=0.083\linewidth]{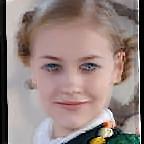} \\
            \includegraphics[width=0.083\linewidth]{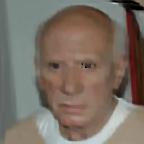}
            \end{tabular}
    }\hspace*{-5.2mm}
    \subfigure[\cite{shen2018deep}]{\begin{tabular}[]{c}
            \includegraphics[width=0.083\linewidth]{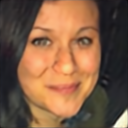} \\
            \includegraphics[width=0.083\linewidth]{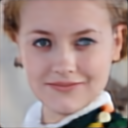} \\
            \includegraphics[width=0.083\linewidth]{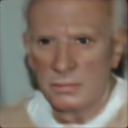}
            \end{tabular}
    }\hspace*{-6.2mm}
    \subfigure[\cite{pan2014deblurring2}]{
        \begin{tabular}[]{c}
            \includegraphics[width=0.083\linewidth]{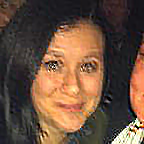} \\
            \includegraphics[width=0.083\linewidth]{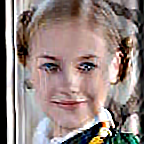} \\
            \includegraphics[width=0.083\linewidth]{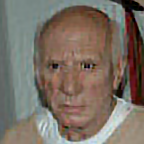}
        \end{tabular}
    }\hspace*{-6mm}
    \subfigure[\cite{xu2013unnatural}]{
        \begin{tabular}[]{c}
            \includegraphics[width=0.083\linewidth]{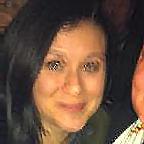} \\
            \includegraphics[width=0.083\linewidth]{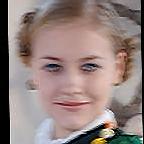} \\
            \includegraphics[width=0.083\linewidth]{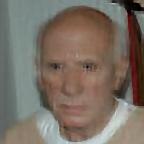}
        \end{tabular}
    }\hspace*{-6.2mm}
    \subfigure[\cite{krishnan2011blind}]{
             \begin{tabular}[]{c}
            \includegraphics[width=0.083\linewidth]{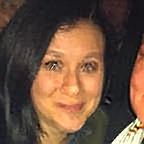} \\
            \includegraphics[width=0.083\linewidth]{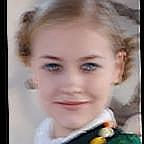} \\
            \includegraphics[width=0.083\linewidth]{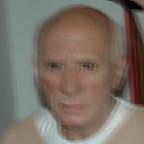}
        \end{tabular}
    }\hspace*{-6.2mm}
    \subfigure[\cite{DeblurGAN}]{
             \begin{tabular}[]{c}
            \includegraphics[width=0.083\linewidth]{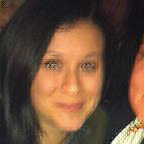} \\
            \includegraphics[width=0.083\linewidth]{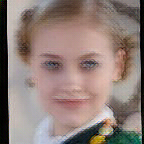} \\
            \includegraphics[width=0.083\linewidth]{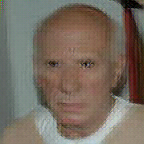}
        \end{tabular}
    }\hspace*{-6.2mm}
    \subfigure[\cite{nah2017deep}]{
            \begin{tabular}[]{c}
            \includegraphics[width=0.083\linewidth]{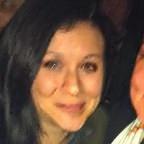} \\
            \includegraphics[width=0.083\linewidth]{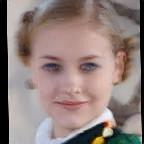} \\
            \includegraphics[width=0.083\linewidth]{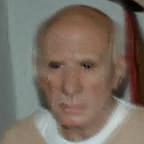}
        \end{tabular}
    }\hspace*{-6mm}
    \subfigure[\cite{CycleGAN2017}]{
            \label{real_cyclegan}
             \begin{tabular}[]{c}
            \includegraphics[width=0.083\linewidth]{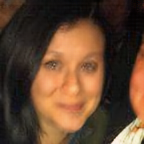} \\
            \includegraphics[width=0.083\linewidth]{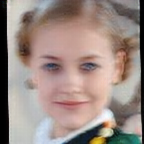} \\
            \includegraphics[width=0.083\linewidth]{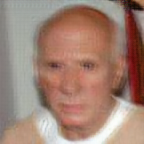}
        \end{tabular}
    }\hspace*{-6.2mm}
    \subfigure[Ours]{
            \label{real_our}
             \begin{tabular}[]{c}
            \includegraphics[width=0.083\linewidth]{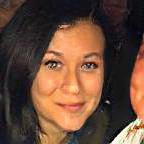} \\
            \includegraphics[width=0.083\linewidth]{figs/real_face/2_10.jpg} \\
            \includegraphics[width=0.083\linewidth]{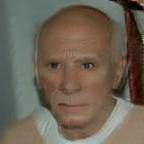}
        \end{tabular}
    }
    \caption{Visual comparisons with state-of-the-art methods on real blurred face images. Best viewed in color and by zooming in.}
    \label{fig:real_face}
    \end{center}
\end{figure*}

\textbf{CFP dataset:} This dataset consists of 7,000 still images from 500 subjects and for each subject, it has ten images in frontal
pose and four images in profile pose. The datasets are divided into ten splits and two protocols: frontal-to-frontal (FF) and frontal-to-profile (FP). We used the same method as described above to blur the images. The faces are detected and aligned similarly as the CelebA dataset.

For CelebA and BMVC\_Text datasets, we use standard debluring metrics (PSNR, SSIM) for evaluation. We also use feature distance (\ie, the $L_2$ distance of the outputs from some deep networks) between the deblurred image and the ground truth image as a measure of semantic similarity because we find this to be a better perceptual metric than PSNR and SSIM~\cite{zhang2018perceptual}. For the CelebA dataset, we use the outputs of $\texttt{pool5}$ layer from VGG-Face~\cite{Parkhi15} and for the text dataset, we use the outputs of $\texttt{pool5}$ layer from a VGG-19 network. For text deblurring, another meaningful metric is the OCR recognition rate for the deblurred text. We follow the same protocol as in~\cite{BMVC2015_6} to report the character error rate (CER) for OCR evaluation. 

To study the influence of motion blur on face recognition and test the performance of different deblurring algorithms, we perform face verification on the CFP dataset. Both frontal-to-frontal and frontal-to-profile protocol are evaluated. The frontal-to-profile protocol can further be used to examine the robustness of the deblurring methods on pose. 

In order to test the generalization capability of the proposed method, we also try our approach on natural images. More details are presented in the supplementary materials.

\begin{table}
\begin{center}
\begin{tabular}{l|c|c|r}
\hline
Method & PSNR & SSIM &$d_{VGG}$\\
\hline
Pan \etal~\cite{pan2014deblurring} & 17.34 & 0.52 & 96.6 \\
Pan \etal~\cite{pan2016blind} & 17.59  & 0.54  & 85.6 \\
Shen \etal~\cite{shen2018deep} &\textbf{21.50} & \textbf{0.69} & \textbf{57.9}\\
Pan \etal~\cite{pan2014deblurring2} & 15.16 & 0.38 & 166.6 \\
Xu \etal~\cite{xu2013unnatural} & 16.84 & 0.47 & 102.0 \\
Krishnan \etal~\cite{krishnan2011blind} & 18.51  & 0.56  & 89.4 \\
Kupyn \etal~\cite{DeblurGAN} & 18.86 & 0.54 & 116.5\\
Nah \etal~\cite{nah2017deep} & 18.26 & 0.57 & 75.6 \\ \hline
Zhu \etal~\cite{CycleGAN2017}  & 19.40 & 0.56 & 103.2 \\
Ours & \textbf{20.81} & \textbf{0.65} & \textbf{57.6} \\
\hline
\end{tabular}
\end{center}
\caption{Quantitative performance comparison with state-of-the-art methods on CelebA dataset. $d_{VGG}$ represents the distance of feature from VGG-Face, lower is better.}
\label{tab:celeba}
\end{table}

\subsection{Ablation Study}
\label{ablation}
In this section, we perform an ablation study to analyze the effectiveness of each component or loss in the proposed framework. Both quantitative and qualitative results on CelebA dataset are reported for the following five variants of our methods where each component is gradually added: 1) only including deblurring branch (\ie, removing the top cycle in Fig.~\ref{fig:system} and the blur encoder $E^b$); 2) adding blurring branch (adding the top cycle of Fig.~\ref{fig:system}); 3) adding content and blur disentanglement; 4) adding the KL divergence loss; 5) adding the perceptual loss. 

\begin{table}[]
\begin{center}
\begin{tabular}{l|c||c}
\hline
Methods & F2F Accuracy & F2P Accuracy \\ \hline
Blurred \etal~\cite{pan2014deblurring} & 0.920$\pm$0.014 & 0.848$\pm$0.013 \\
Sharp \etal~\cite{pan2014deblurring} & 0.988$\pm$0.005 & 0.949$\pm$0.014 \\ \hline
Pan \etal~\cite{pan2014deblurring} & 0.930$\pm$0.013 & 0.853$\pm$0.010          \\ 
Pan \etal~\cite{pan2016blind} & 0.935$\pm$0.015 & 0.872$\pm$0.015   \\
Shen \etal~\cite{shen2018deep} & 0.959$\pm$0.008 & 0.821$\pm$0.022 \\
Pan \etal~\cite{pan2014deblurring2} & 0.916$\pm$0.011 & 0.825$\pm$0.016 \\
Xu \etal~\cite{xu2013unnatural} & 0.944$\pm$0.012& 0.865$\pm$0.013 \\
Krishnan \etal~\cite{krishnan2011blind} & 0.941$\pm$0.012& 0.857$\pm$0.014\\
Kupyn \etal~\cite{DeblurGAN} & 0.948$\pm$0.012& 0.872$\pm$0.007 \\
Nah \etal~\cite{nah2017deep} & \textbf{0.960$\pm$0.007} & \textbf{0.885$\pm$0.016} \\  \hline
Zhu \etal~\cite{CycleGAN2017}  & 0.941$\pm$0.012& 0.864$\pm$0.015 \\
Ours &  \textbf{0.948$\pm$0.006} & \textbf{0.872$\pm$0.015} \\ \hline
\end{tabular}
\end{center}
\caption{Face verification results on the CFP dataset. F2F, F2P represent frontal-to-frontal and frontal-to-profile protocols. }
\label{tab:cfp}
\end{table}

\begin{figure*}
    \begin{center}
    \hspace*{-1mm}
    \subfigure[Blurred]{ \begin{tabular}[]{c}
        \includegraphics[width=0.12\linewidth]{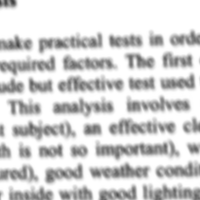} \\
        \includegraphics[width=0.12\linewidth]{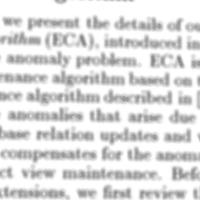}
            \end{tabular}
    }\hspace*{-5.1mm}
    \subfigure[\cite{pan2014deblurring}]{\begin{tabular}[]{c}
            \includegraphics[width=0.12\linewidth]{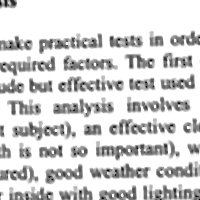} \\
            \includegraphics[width=0.12\linewidth]{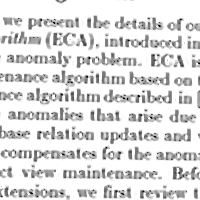}
            \end{tabular}
    }\hspace*{-5.4mm}
    \subfigure[\cite{pan2016blind}]{\begin{tabular}[]{c}
            \includegraphics[width=0.12\linewidth]{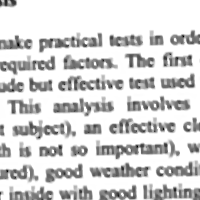} \\
            \includegraphics[width=0.12\linewidth]{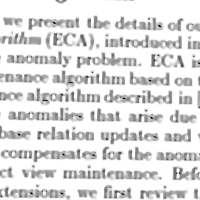}
            \end{tabular}
    }\hspace*{-5.2mm}
    \subfigure[\cite{nah2017deep}]{\begin{tabular}[]{c}
            \includegraphics[width=0.12\linewidth]{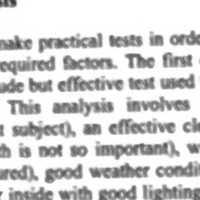} \\
            \includegraphics[width=0.12\linewidth]{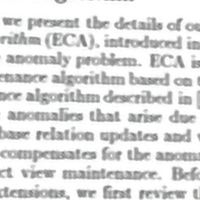}
            \end{tabular}
    }\hspace*{-6.2mm}
    \subfigure[\cite{CycleGAN2017}]{
        \begin{tabular}[]{c}
            \includegraphics[width=0.12\linewidth]{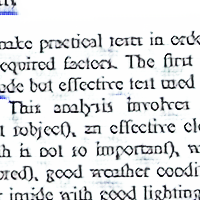} \\
            \includegraphics[width=0.12\linewidth]{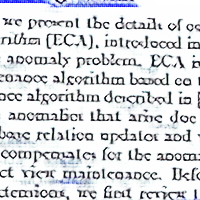}
        \end{tabular}
    }\hspace*{-6mm}
    \subfigure[\cite{BMVC2015_6}]{
        \begin{tabular}[]{c}
            \includegraphics[width=0.12\linewidth]{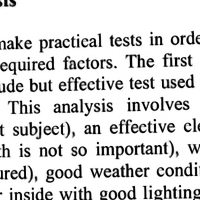} \\
            \includegraphics[width=0.12\linewidth]{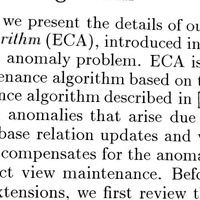}
        \end{tabular}
    }\hspace*{-6.2mm}
    \subfigure[Ours]{
            \label{fig:text_ours}
             \begin{tabular}[]{c}
            \includegraphics[width=0.12\linewidth]{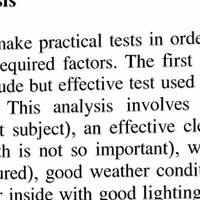} \\
            \includegraphics[width=0.12\linewidth]{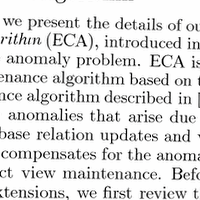}
        \end{tabular}
    }\hspace*{-6.2mm}
    \subfigure[Sharp]{
            \label{fig:text_sharp}
            \begin{tabular}[]{c}
            \includegraphics[width=0.12\linewidth]{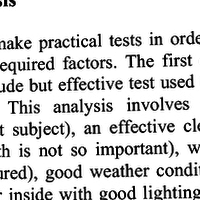} \\
            \includegraphics[width=0.12\linewidth]{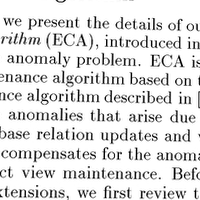}
        \end{tabular}
    }
    \caption{Visual results compared with state-of-the-art methods on BMVC\_Text dataset. Best viewed by zooming in.}
    \label{fig:text}
    \end{center}
\end{figure*}

\begin{figure*}
    \begin{center}
    \hspace*{-6mm}
    \subfigure[Blurred]{ \begin{tabular}[]{c}
        \includegraphics[width=0.136\linewidth]{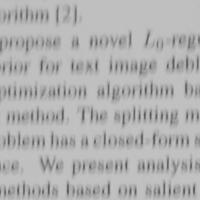} \\
        \includegraphics[width=0.136\linewidth]{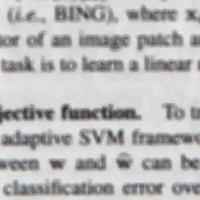}
            \end{tabular}
    }\hspace*{-5.1mm}
    \subfigure[\cite{pan2014deblurring}]{\begin{tabular}[]{c}
            \includegraphics[width=0.136\linewidth]{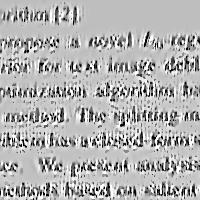} \\
            \includegraphics[width=0.136\linewidth]{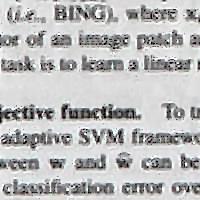}
            \end{tabular}
    }\hspace*{-5.4mm}
    \subfigure[\cite{pan2016blind}]{\begin{tabular}[]{c}
            \includegraphics[width=0.136\linewidth]{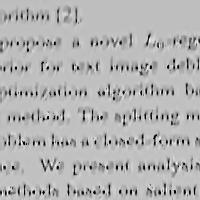} \\
            \includegraphics[width=0.136\linewidth]{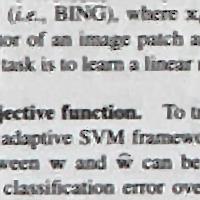}
            \end{tabular}
    }\hspace*{-5.2mm}
    \subfigure[\cite{nah2017deep}]{\begin{tabular}[]{c}
            \includegraphics[width=0.136\linewidth]{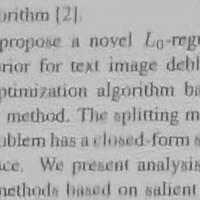} \\
            \includegraphics[width=0.136\linewidth]{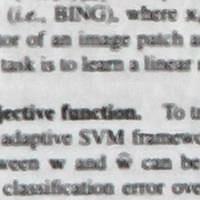}
            \end{tabular}
    }\hspace*{-6.2mm}
    \subfigure[\cite{CycleGAN2017}]{
        \begin{tabular}[]{c}
            \includegraphics[width=0.136\linewidth]{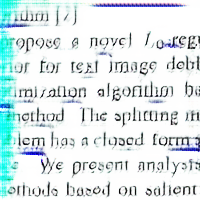} \\
            \includegraphics[width=0.136\linewidth]{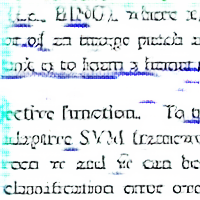}
        \end{tabular}
    }\hspace*{-6mm}
    \subfigure[\cite{BMVC2015_6}]{
        \begin{tabular}[]{c}
            \includegraphics[width=0.136\linewidth]{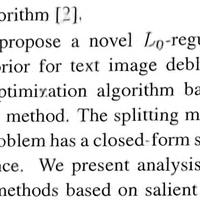} \\
            \includegraphics[width=0.136\linewidth]{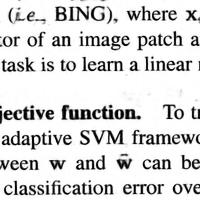}
        \end{tabular}
    }\hspace*{-6.2mm}
    \subfigure[Ours]{
            \label{fig:realtext_ours}
             \begin{tabular}[]{c}
            \includegraphics[width=0.13\linewidth]{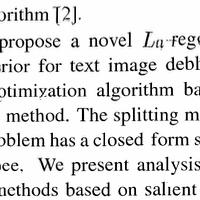} \\
            \includegraphics[width=0.13\linewidth]{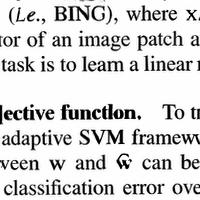}
        \end{tabular}
    }\hspace*{-6.2mm}
    \caption{Visual results compared with state-of-the-art methods on real blurred text images. Best viewed by zooming in.}
    \label{fig:real_text}
    \end{center}
\end{figure*}

We present the PSNR, SSIM and VGG-Face distance ($d_{VGG}$) for each variant in Table~\ref{tab:ablation} and the visual comparisons are shown in Fig.~\ref{fig:ablation}. From Table~\ref{tab:ablation}, we can see that adding the blurring branch significantly improves the deblurring performance, especially for the perceptual distance. As shown in Fig.~\ref{fig:ablation} (c) many artifacts are removed from face and colors are preserved well compared to (b). This confirms the findings in CycleGAN~\cite{CycleGAN2017} that only one direction cycle-consistency loss is not enough to recover good images. However, we find that adding a disentanglement component does not help but rather hurt the performance ( Fig.~\ref{fig:ablation} (d)). This demonstrates that the blurring encoder $E^b$ will induce some noise and confuse the generator $G_S$ if the KL divergence loss is not enforced. In contrast, when the KL diveregence loss is added to $E^b$ (Fig.~\ref{fig:ablation} (e)), content and blur information can be better disentangled and we observe some improvements on both PSNR and perceptual similarities. Finally, the perceptual loss can improve the perceptual reality of the face notably. By comparing Fig.~\ref{fig:ablation} (e) and (f), we find that the artifacts on cheek and forehead are further removed. Furthermore, the mouth region of (f) is more realistic than (e).

\subsection{Face Results}
\label{exp_face}
\textbf{Compared methods:} We compare the proposed method with some state-of-the-art deblurring methods~\cite{pan2014deblurring,pan2016blind,shen2018deep,pan2014deblurring2,xu2013unnatural,krishnan2011blind,nah2017deep,CycleGAN2017,DeblurGAN}. We directly use the pre-trained models provided by authors except for CycleGAN~\cite{CycleGAN2017}, where we retrain the model by using the same training set as our method. Both CNN-based models ~\cite{shen2018deep,nah2017deep,CycleGAN2017,DeblurGAN} and conventional MAP-based methods are included~\cite{pan2014deblurring,pan2016blind,pan2014deblurring2,xu2013unnatural,krishnan2011blind}. Among these approaches, two are specific for face deblurring~\cite{pan2014deblurring2,shen2018deep} while others are generic deblurring algorithm. The kernel size for~\cite{pan2014deblurring,pan2016blind} is set to 9. We found that the face deblurring method~\cite{shen2018deep} is very sensitive to face alignment, we follow the sample image provided by the author to align the faces before running their algorithm. Meanwhile, CycleGAN is the only unsupervised CNN-based method we compare with. 

\textbf{CelebA dataset results.} The quantitative results for CelebA dataset are shown in Table~\ref{tab:celeba} and the visual comparisons are illustrated in Fig.~\ref{fig:celeba}. Our approach shows superior performance to other unsupervised algorithms on both conventional metrics and VGG-Face distance. Furthermore, we achieve comparable results with state-of-the-art supervised face deblurring method~\cite{shen2018deep}. From Fig.~\ref{fig:celeba} we can see that conventional methods often over-deblur or under-deblur the blurred images. Among them, Krishnan \etal~\cite{krishnan2011blind} perform the best in PSNR and SSIM and Pan \etal~\cite{pan2016blind} perform the best in perceptual distance. For CNN-based methods, Shen \etal~\cite{shen2018deep} include a face parsing branch and achieve the best performance among the compared methods. The results for DeblurGAN~\cite{DeblurGAN} contain some ringing artifacts and CycleGAN~\cite{CycleGAN2017} cannot recover the mouth part of both images that well. Nah \etal~\cite{nah2017deep} shows better visual results than other CNN-based generic methods but still contains some blur in local structures. 


\textbf{Face verification results.} The face verification results for the CFP dataset are reported in Table~\ref{tab:cfp}. We train a 27-layer ResNet~\cite{lu2019experimental} on the curated MS-Celeb1M dataset~\cite{guo2016msceleb,lin2017proximity} with 3.7 millions face images and extract features of the deblurred faces for each method. Cosine similarities of test pairs are used as similarity scores for face verification. We follow the protocols used in~\cite{lu2016regularized,lu2017pose} and the verification accuracy for both frontal-to-frontal and frontal-to-profile protocols are reported. As shown in Table~\ref{tab:cfp}, the proposed method improves the baseline results of blurred images and outperforms CycleGAN~\cite{CycleGAN2017} on both protocols. Moreover, we achieve comparable performance compared to other state-of-the-art supervised deblurring methods. Shen \etal~\cite{shen2018deep} perform very well for frontal-to-frontal protocol, yet provide the worst performance on frontal-to-profile protocol, which shows that the face parsing network in their method is sensitive to poses. In contrast, the proposed method works for both frontal and profile face images even though we do not explicitly train on faces with extreme poses. 

\begin{table}
\begin{center}
\begin{tabular}{l|c|c|c|c}
\hline
Method & PSNR & SSIM &$d_{VGG}$ & CER\\
\hline
Pan \etal~\cite{pan2014deblurring} & 21.18 & 0.92 & 19.7 & 42.3\\
Pan \etal~\cite{pan2016blind} & 21.84  & 0.93  & 15.7 & 35.3\\
Nah \etal~\cite{nah2017deep} & 22.27 & 0.92 & 31.9 & 50.6\\
Hradis \etal~\cite{BMVC2015_6} & \textbf{30.6} & \textbf{0.98} & \textbf{1.6} & \textbf{7.2}\\ \hline
Zhu \etal~\cite{CycleGAN2017}  & 19.57 & 0.89 & 18.8 & 53.0\\
Ours & \textbf{22.56} & \textbf{0.95} & \textbf{2.2} & \textbf{10.1}\\
\hline
\end{tabular}
\end{center}
\caption{Quantitative performance comparison with state-of-the-art methods on BMVC\_Text dataset. $d_{VGG}$ represents the distance of feature from VGG-Face, lower is better. CER is the OCR character error rate, lower is better.}
\label{tab:text}
\end{table}
\textbf{Real blurred images results}
We also evaluate the proposed method on some real-world images from the datasets of Lai \etal~\cite{lai2016comparative}, and the results are shown in Fig.~\ref{fig:real_face}. Similar to what we have observed for CelebA, our method shows competitive performance compared to other state-of-the-art approaches. Conventional methods~\cite{pan2014deblurring,pan2016blind,pan2014deblurring2,xu2013unnatural,krishnan2011blind} still tend to under-deblur or over-deblur images, especially on local regions such as eyes and mouths. On the other hand, the generic CNN-based method~\cite{DeblurGAN} does not perform very well on face deblurring. CycleGAN~\cite{CycleGAN2017} fails to recover sharp faces but only changes the background color of images (\eg, third row of Fig.~\ref{real_cyclegan}). Nah \etal~\cite{nah2017deep} produce good results on the first two faces, but generate some artifacts in the third image. Deep semantic face deblurring~\cite{shen2018deep} generate better results than other compared methods. Nonetheless, due to the existence of face parsing, they tend to sharpen some facial parts (eye, nose and mouth) but over-smooth the ears and the background. In contrast, our method can not only recover sharp faces, but also restore sharp textures in the background (\eg, third row of Fig.~\ref{real_our}).

\subsection{Text results}
\label{exp_text}
\textbf{BMVC\_Text dataset results.} 
Similar to face experiments, we train a CycleGAN model using the same training set as our method. The kernel size for~\cite{pan2014deblurring,pan2016blind} is set to 12. The quantative results for BMVC\_Text dataset are shown in Table~\ref{tab:text} and some sample images are presented in Fig.~\ref{fig:text}. We can see that conventional methods~\cite{pan2014deblurring,pan2016blind} and generic deblurring approaches~\cite{nah2017deep} do not perform well on text deblurring. The visual quality is poor and the OCR error rate is very high. The results for CycleGAN~\cite{CycleGAN2017} contain some weird blue background. Although it removes the blur in images, it fails to recover recognizable text. In contrast, our method achieves good visual quality and its performance is comparable to the state-of-the-art supervised text deblurring method~\cite{BMVC2015_6} on semantic metrics (\ie, perceptual distance and OCR error rate). Interestingly, we find the PNSR performance for our approach is worse than the method~\cite{BMVC2015_6} by large margins. We carefully examine our visual results and find that the proposed method sometimes changes the font of the text while deblurring. For example, as shown in the first row of Fig.~\ref{fig:text_ours}, the font of our deblurred text becomes lighter and thinner compared to the original sharp text image (Fig.~\ref{fig:text_sharp}). The main reason for this phenomenon is that our method does not utilize paired training data so that the deblurring generator cannot preserve some local details of text images.

\textbf{Real blurred text images results} We also evaluate our deblurring method on real blurred text images provided by Hradis \etal~\cite{BMVC2015_6}. Due to space limitation, $200 \times 200$ patches are randomly cropped, and some visual results are illustrated in Fig.~\ref{fig:real_text}. Similar to the results of BMVC\_Text dataset, we find that conventional methods~\cite{pan2014deblurring,pan2016blind} fail to deblur the given text images. Nah \etal~\cite{nah2017deep}, in contrast, generate a reasonable deblurred result for the first image but cannot handle the second one. CycleGAN~\cite{CycleGAN2017} again produces blue artifacts and cannot recover meaningful text information. Hradis\etal~\cite{BMVC2015_6} and our approach both generate satisfactory results. Although we mis-recognize some characters (\eg, in the second images, "\ie, BING" is recovered as "\textit{Le.},BING"), we still correctly recover most of the blurred images.

\section{Conclusions}

In this paper, we propose an unsupervised method for domain-specific single image deblurring. We disentangle the content and blur features in a blurred image and add the KL divergence loss to discourage the blur features to encode content information. In order to preserve the content structure of the original images, we add a blurring branch and cycle-consistency loss to the framework. The perceptual loss helps the blurred image remove unrealistic artifacts. Ablation study on each component shows the effectiveness of different modules. We conduct extensive experiments on face and text deblurring. Both quantative and visual results show promising performance compared to other state-of-the-art approaches.

\section{Acknowledgment}
\small{
This research is based upon work supported by the Office of the Director of National Intelligence (ODNI), Intelligence Advanced Research Projects Activity (IARPA), via IARPA R\&D Contract No. 2014-14071600012. The views and conclusions contained herein are
those of the authors and should not be interpreted as necessarily representing the official policies or endorsements, either expressed or implied, of the ODNI, IARPA, or the U.S. Government. The U.S. Government is authorized to reproduce and distribute reprints for
Governmental purposes notwithstanding any copyright annotation thereon.
}
{\small
\bibliographystyle{ieee}
\bibliography{egbib}
}

\end{document}